\documentclass[a4paper]{article}

\usepackage{INTERSPEECH2018}

\usepackage{algorithm}
\usepackage{algorithmicx}
\usepackage{multirow}
\usepackage[noend]{algpseudocode}
\usepackage{xcolor}
\usepackage[font=footnotesize,labelfont=bf]{caption}
\usepackage{booktabs}
\usepackage[pagebackref=true,breaklinks=true,letterpaper=true,colorlinks,bookmarks=false]{hyperref}
\usepackage{diagbox} 

\title{ LRS3-TED: a large-scale dataset for visual speech recognition}
\name{Triantafyllos Afouras, Joon Son Chung, Andrew Zisserman}
\address{
  Visual Geometry Group, Department of Engineering Science,\\
  University of Oxford, UK}
  \email{\{afourast, joon, az\}@robots.ox.ac.uk}

\begin{document}

\maketitle
\begin{abstract}

This paper introduces a new multi-modal dataset for 
visual and audio-visual speech recognition. 
It includes face tracks from over 400 hours
of TED and TEDx videos, along with the corresponding 
subtitles and word alignment boundaries. 
The new dataset is substantially larger in scale compared to 
other public datasets that are available for general research.

\end{abstract}


\section{Introduction}

Visual speech recognition (or {\em lip reading}) is a very challenging task,
and a difficult skill for a human to learn. 
In the recent years, there has been significant progress~\cite{Chung17,Assael16,Stafylakis17,shillingford2018large} in the performance of automated lip
reading due to the application of deep neural network models and the availability of large scale
datasets.
However, most of these 
datasets are subject to some restrictions (e.g.\ LRW~\cite{Chung16} or the 
LRS2-BBC~\cite{Afouras18c} cannot be used by industrial research labs) and this has
   meant that it is difficult to compare the performance of one 
lip-reading system to another, as there is no large scale common
   benchmark dataset. Our aim in releasing the LRS3-TED dataset is to 
provide such a benchmark dataset, and one that is larger in size
   compared to any available dataset in this field.
   
   The \texttt{LRS3-TED} dataset
 can be downloaded from {\scriptsize\url{http://www.robots.ox.ac.uk/~vgg/data/lip_reading}}.
   


\begin{table*}[t!]
\setlength{\tabcolsep}{9pt}
\centering
\footnotesize
\begin{tabular}{ l  c  l  c  r   r  r  r r }
  \toprule
  \textbf{Dataset} & \textbf{Source} & \textbf{Split} & \textbf{Dates} & \textbf{\# Spk.} &
  \textbf{\# Utt.} & \textbf{Word inst.} & \textbf{Vocab} & \textbf{\# hours}  \\ 
  \midrule 
  
   GRID~\cite{Cooke06} & - & - & - 					& 51 & 33,000 & 165k & 51 & 27.5\\ \midrule
   MODALITY~\cite{Czyzewski17} & - & - & - 		& 35 &  5,880 & 8,085 & 182 & 31 \\ 
   \midrule
   
  \multirow{2}{*}{LRW~\cite{Chung16}}  &  \multirow{2}{*}{BBC}  & Train-val & 01/2010 - 12/2015 & -
  & 514k & 514k & 500 & 165 \\ 
   &  & Test & 01/2016 - 09/2016 & - & 25k & 25k & 500  & 8\\
   \midrule 

 \multirow{4}{*}{LRS2-BBC~\cite{Afouras18c}} & \multirow{4}{*}{BBC} 
 &   Pre-train  & 01/2010 - 02/2016   &  - & 96k & 2M & 41k & 195 \\ 
  &  &  Train-val  & 01/2010 - 02/2016    &  - & 47k & 337k & 18k & 29 \\ 
  &  & Test        & 03/2016 - 09/2016   &  - & 1,243 &  6,663 & 1,693 & 0.5 \\ 
  &  & Text-only        & 01/2016 - 02/2016 &  -  &  8M & 26M   & 60k & - \\ 
  \midrule

 \multirow{4}{*}{{\bf LRS3-TED}} &   \multirow{4}{*}{\shortstack{TED \& \\ TEDx\\(YouTube)}} 
 &   Pre-train  & -  & 5,090 & 119k & 3.9M & 51k & 407 \\ 
  & &  Train-val                                       & -   & 4,004 & 32k & 358k & 17k & 30  \\ 
  & & Test                                               & - & 451 & 1,452 &  11k & 2,136 & 1  \\ 
  &  & Text-only        &  - &  5,543 & 1.2M  &  7.2M & 57k & - \\ 
  \midrule 

\end{tabular} 
\normalsize
\caption{A comparison of publicly available lip reading datasets. Division of training, validation and test data; and the number of utterances, number of word instances and vocabulary size of each partition.
{\bf Utt:} Utterances.
}
\label{table:datastat}
\end{table*}


\section{LRS3-TED dataset}

The dataset consists of over 400 hours of video, extracted from 5594 TED and TEDx talks in English, downloaded from YouTube.

The cropped face tracks are provided as \texttt{.mp4} files with a resolution of
224$\times$224 and a frame rate of 25 fps, 
encoded using the h264 codec. 
The audio tracks are provided as single-channel 16-bit 16kHz format, while 
the corresponding text transcripts, as well as the alignment boundaries of every word are included
in plain text files.

The dataset is organized into three  sets: {\em pre-train}, {\em train-val} and {\em test}. The first two
overlap in terms of content but the last is completely independent. The statistics for each set are given in Table~\ref{table:datastat}.

\subsection{Data collection}


We use a multi-stage pipeline for 
automatically generating the large-scale dataset for
audio-visual speech recognition. 
Using this pipeline, we have been
able to collect hundreds of hours of spoken sentences and
phrases along with the corresponding facetrack.

We start from the TED and TEDx videos that are available on their
respective YouTube channels.
These videos were selected for mutliple reasons: 
(1) a wide range of speakers appears
in the videos, unlike movies or dramas
with a fixed cast; 
(2) shot changes are less frequent,
therefore there are more full sentences with 
continuous facetracks;
(3) the speakers usually talk without interruption,
allowing us to obtain longer face tracks.
TED videos have previously been used for audio-visual datasets for these 
reasons~\cite{Ephrat18}.

The pipeline is based on the methods
described in~\cite{Chung17, Afouras18c},
but we give a brief sketch of the method here.

\noindent\textbf{Video preparation.} 
We use a CNN face detector based on the
Single Shot MultiBox Detector (SSD)~\cite{Liu16}
to detect face appearances in the individual frames.

The time boundaries of a shot are determined by comparing color histograms
across consecutive frames~\cite{Lienhart01}, and
within each shot, face tracks are generated from face detections
based on their positions.

\noindent\textbf{Audio and text preparation.}
Only the videos providing english subtitles created by humans were used.
The subtitles in the YouTube videos are broadcast
in sync with the audio only at sentence-level,
therefore
the Penn Phonetics Lab Forced
Aligner (P2FA)~\cite{Yuan08} 
is used to obtain
a word-level alignment between the subtitle
and the audio signal. 
The alignment is double-checked
against an off-the-shelf Kaldi-based ASR model.

\noindent\textbf{AV sync and speaker detection.}
In YouTube or broadcast videos, 
the audio and the video streams can be out of
sync by up to around one second, which can introduce temporal
offsets between the videos and the text labels (aligned to the audio).
We use a two-stream network (SyncNet) described in \cite{Chung16a} to 
synchronise the two streams. 
The same network is also used to determine 
which face's lip movements match the audio,
and if none matches, the clip is rejected
as being a voice-over.

\noindent\textbf{Sentence extraction.} 
The videos are divided into individual sentences/ phrases using the punctuations
in the transcript. 
The sentences
are separated by full stops, commas 
and question marks.
The sentences in the {\em train-val} and {\em test}  sets
are clipped to 100 characters or 6 seconds. 

The {\em train-val} and {\em test} sets are divided by videos (extracted from disjoint sets of original videos).
Although we do not explicitly label the identities,
it is unlikely that there are many identities that appear in
both training and test sets, since the speakers do not
generally appear on TED programs repeatedly.
This is in contrast to the LRW and the LRS2-BBC
datasets that are based on regular TV programs, hence
the same characters are likely to appear in common
from one episode to the next.

The {\em pre-train} set is more extensive, as it contains 
videos spanning the full duration of the face track, along with the corresponding subtitles.
It is extracted from the same set of original YouTube videos as the {\em train-val} set.
However, these videos may be shorter or longer than the full sentences included in the {\em
train-val} and {\em test} sets,
and are annotated with the alignment boundaries of every word.


\section{Conclusion}

In this document, we have briefly described the LRS3-TED audio-visual corpus. 
The dataset is useful for many applications including lip reading, audio-visual speech
recognition, video-driven speech enhancement, as well as other audio-visual learning tasks.
\cite{Afouras18c} reports the performance of some of the latest lip reading models on this dataset.


\vspace{10pt}

\noindent\textbf{Acknowledgements.}
Funding for this research is provided by the UK EPSRC
CDT in Autonomous Intelligent Machines and Systems, 
the Oxford-Google DeepMind Graduate Scholarship, and by the EPSRC 
Programme Grant Seebibyte EP/M013774/1. 


\bibliographystyle{IEEEtran}

\bibliography{shortstrings,vgg_other,mybib,vgg_local}

\clearpage

\end{document}